\documentclass[10pt,twocolumn,letterpaper]{article}

\usepackage{cvpr}              
\usepackage{colortbl}
\usepackage{xcolor}
\usepackage[accsupp]{axessibility}
\usepackage{makecell}
\newcommand{\Diff}[1]{\,\mathrm{d}#1}
\newcommand{\model}{\textit{TensoFlow}}
\usepackage{float}

\definecolor{cvprblue}{rgb}{0.21,0.49,0.74}
\usepackage[pagebackref,breaklinks,colorlinks,allcolors=cvprblue]{hyperref}

\def\paperID{2509} 
\def\confName{CVPR}
\def\confYear{2025}

\title{TensoFlow: Tensorial Flow-based Sampler for Inverse Rendering}

\author{%
  Chun Gu$^{1,2}$
  \quad 
  Xiaofei Wei$^{1,2}$
  \quad
  Li Zhang$^{1,2}$\thanks{Corresponding author (\url{lizhangfd@fudan.edu.cn}).}
  \quad 
  Xiatian Zhu$^3$ 
  \\
  $^1$School of Data Science, Fudan University 
  \quad
  $^2$Shanghai Innovation Institute
  \quad
  $^3$University of Surrey 
  \vspace{.5em} 
  \\
  \textcolor{black}{\url{https://github.com/fudan-zvg/tensoflow}}
}

\begin{document}

\maketitle

\begin{abstract}

Inverse rendering aims to recover scene geometry, material properties, and lighting from multi-view images. Given the complexity of light-surface interactions, importance sampling is essential for the evaluation of the rendering equation, as it reduces variance and enhances the efficiency of Monte Carlo sampling. Existing inverse rendering methods typically use pre-defined non-learnable importance samplers in prior manually, struggling to effectively match the spatially and directionally varied integrand and resulting in high variance and suboptimal performance. To address this limitation, we propose the concept of learning a spatially and directionally aware importance sampler for the rendering equation to accurately and flexibly capture the unconstrained complexity of a typical scene. We further formulate \textbf{\model{}}, a generic approach for sampler learning in inverse rendering, enabling to closely match the integrand of the rendering equation spatially and directionally. Concretely, our sampler is parameterized by normalizing flows, allowing both directional sampling of incident light and probability density function (PDF) inference. To capture the characteristics of the sampler spatially, we learn a tensorial representation of the scene space, which imposes spatial conditions, together with reflected direction, leading to spatially and directionally aware sampling distributions. Our model can be optimized by minimizing the difference between the integrand and our normalizing flow. Extensive experiments validate the superiority of \model{} over prior alternatives on both synthetic and real-world benchmarks.

\end{abstract}    
\section{Introduction}
\label{sec:intro}

3D scene modeling has made significant strides, especially with the emergence of powerful representation models like neural radiance fields (NeRF) \cite{mildenhall2021nerf,barron2021mip,barron2022mip,barron2023zipnerf} and efficient ones like 3D Gaussian splatting (3DGS) \cite{kerbl3Dgaussians,huang20242d,yu2024mip}.
However, these methods struggle with inverse rendering due to the inability to effectively disentangle geometry, material properties, and lighting, restricting their capacity to relight scenes.

To address this issue, recent works \cite{yao2022neilf,liu2023nero,jin2023tensoir,li2024tensosdf,gao2023relightable,liang2024gs,jiang2024gaussianshader,yao2025refGS,gu2024irgs} have advanced appearance modeling by incorporating the physically-based rendering equation~\cite{kajiya1986rendering} that simulates light-surface interactions. This however introduces a fundamental challenge -- a need for accurately evaluating the rendering equation using Monte Carlo sampling. One intuitive method is stratified sampling over the hemisphere \cite{yao2022neilf,jin2023tensoir}, but this can be computationally inefficient due to the need for many samples.
To improve efficiency, leading methods like NeRO \cite{liu2023nero} and TensoSDF \cite{li2024tensosdf} employ importance sampling. These approaches utilize pre-defined importance samplers (\eg, cosine-weighted and GGX samplers~\cite{walter2007microfacet} for diffuse and specular components, respectively). However, these pre-defined importance samplers still result in high variance, as the integrand typically varies in both space and direction, making it difficult for a fixed sampler to effectively match the scene’s characteristics.

To address this challenge, we propose a novel perspective of learning spatially and directionally aware importance samplers, allowing us to manage the complexity of any scene without the need for manual sampler selection, while benefiting from greater flexibility and generality. As a showcase, we formulate \model{}, a concrete approach for importance sampler learning. Our sampler is parameterized by normalizing flows composed with piecewise-quadratic coupling layers~\cite{muller2019neural}, functioning as a neural distribution that enables both
probability density function (PDF) inference and directional sampling of incident light in the rendering equation.
To effectively capture the characteristics of the sampler spatially, we learn a tensorial representation of the scene space, imposing spatial conditions on the rendering process, together with reflected direction, resulting in spatially and directionally aware sampling distributions. This formulation can be optimized by minimizing the distributional difference between the integrand and our normalizing flow.

We make these {\bf contributions}:
\textbf{(i)}
We propose a novel concept of learning the importance samplers of the rendering equation for inverse rendering, in contrast to existing pre-defined non-learnable samplers struggling intrinsically to capture the intrinsic rendering complexity of a scene. 
\textbf{(ii)}
We introduce an effective method for sampler learning, \model{}, parameterized by tensorial normalizing flows, serving as a neural distribution that can support both directional sampling of incident light and PDF inference required in the rendering equation.
\textbf{(iii)}
Extensive experiments on both synthetic and real-world datasets demonstrate that \model{} outperforms previous alternatives, achieving significantly lower variance in evaluating the rendering equation with the same number of sampled directions.

\section{Related work}
\begin{figure*}[ht]
\centering
\includegraphics[width=\linewidth]{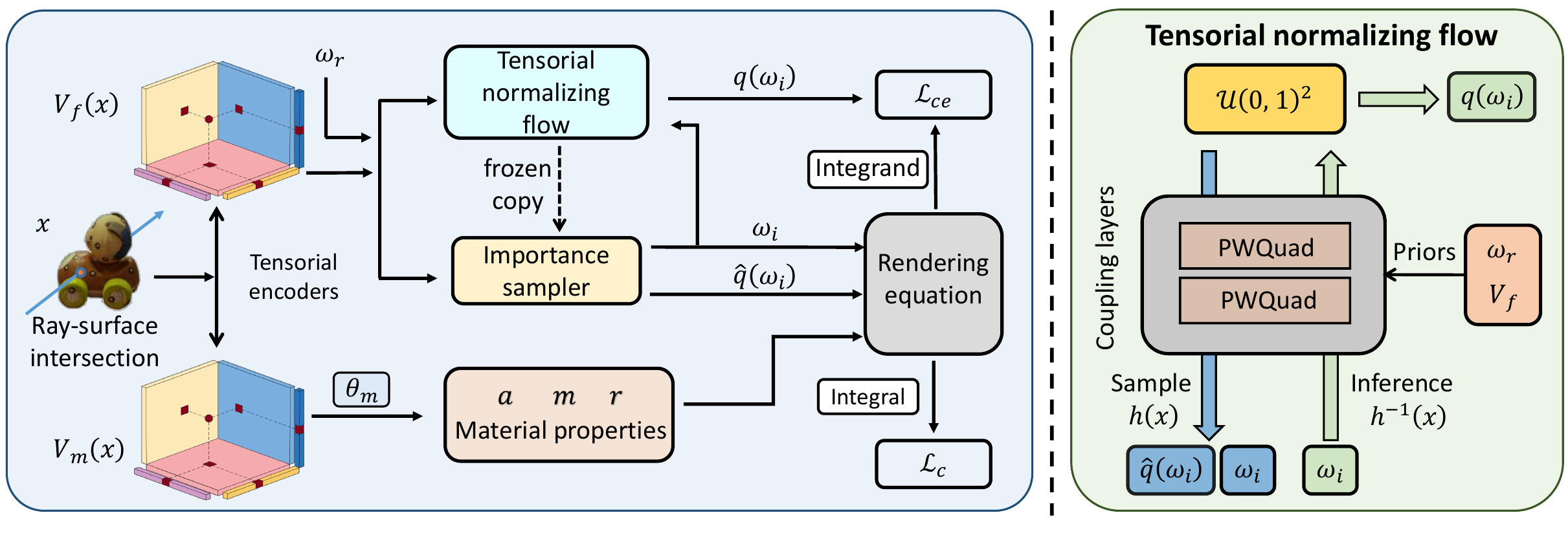}
\caption{
\textbf{Left: Material and lighting estimation stage of \model{}.} Given a ray-surface intersection point $\boldsymbol{x}$, we use tensorial encoders to encode latent features for both the material properties and the importance sampler. When evaluating the rendering equation, incident directions are sampled from our learnable importance sampler, which is a frozen copy of the training normalizing flow. The normalizing flow is optimized by minimizing the distribution difference $\mathcal{L}_\mathrm{ce}$ between $q(\boldsymbol{\omega}_i)$ and the integrand. The material properties, parameterized by $\theta_m$, are optimized by minimizing the RGB rendering loss $\mathcal{L}_\mathrm{c}$ of the final rendering integral.
\textbf{Right: Tensorial normalizing flow.} With spatial prior $V_f$ and directional prior $\boldsymbol{\omega}_r$, our tensorial normalizing flow, implemented using piecewise-quadratic coupling layers, enables both incident direction sampling and PDF querying.
}
\label{fig:pipeline}
\end{figure*}

\paragraph{Novel view synthesis}
The introduction of neural radiance fields (NeRF)~\cite{mildenhall2021nerf}, which represent 3D scenes as continuous volumetric functions, utilizes volumetric rendering techniques to synthesize images from novel viewpoints, achieving effective 3D scene reconstruction. However, NeRF exhibits limitations in terms of rendering efficiency. Various efficient alternatives have been proposed, including voxel-based methods~\cite{sun2022direct,yu_and_fridovichkeil2021plenoxels}, multi-resolution hash grids~\cite{muller2022instant}, and tensor decomposition approaches~\cite{chen2022tensorf}.
Additionally, techniques such as anti-aliasing~\cite{barron2021mip,barron2022mip,barron2023zipnerf} and modeling with signed distance fields (SDFs)~\cite{wang2021neus,neus2,yariv2021volume,wu2022voxurf,li2023neuralangelo} have been employed to enhance rendering and geometry reconstruction quality. Recently, 3D Gaussian splatting (3DGS)~\cite{kerbl3Dgaussians} has been introduced with favored efficiency, representing 3D scenes as a collection of 3D Gaussians. However, despite its impressive performance in rendering quality and efficiency, 3DGS lacks robust geometry, which limits its effectiveness in geometry-sensitive tasks such as inverse rendering~\cite{gao2023relightable,liang2024gs}.

\paragraph{Inverse rendering}

Inverse rendering aims to recover the 3D structure, lighting conditions, and material properties of a scene from posed 2D images. 
Recent advancements in neural rendering have substantially driven inverse rendering performance. NeRF-based approaches~\cite{zhang2021nerfactor,yao2022neilf,zhang2023neilf++,boss2021nerd,knodt2021neural,jin2023tensoir,liu2023nero,li2024tensosdf,attal2024flash} simulate physical light-surface interactions, allowing for the disentanglement of scene geometry, materials, and lighting from image observations. 
In particular,
Monte Carlo sampling is commonly used to evaluate the rendering equation in inverse rendering. 
For example, avoiding ray marching on incident rays, NeILF~\cite{yao2022neilf} optimizes an incident light field using stratified sampling (\eg, sampling directions uniformly over the hemisphere) for the rendering equation.
Similarly, TensoIR~\cite{jin2023tensoir} also applies stratified sampling during training using tensorial implicit fields. 
However, stratified sampling can be computationally costly due to the high sample count needed for accuracy. 
To address this issue, NeRO~\cite{liu2023nero} and TensoSDF~\cite{li2024tensosdf} 
adopt importance sampling with pre-defined samplers (\eg, cosine-weighted, GGX).
Commonly, all these sampling methods fall short of matching the spatially and directionally varying integrand of the rendering equation, leading to suboptimal rendering performance.
To overcome this limitation, we introduce the concept of learning importance samplers
that addressing the need for spatially and directionally aware samplers elegantly.

\section{Preliminary}\label{sec:preliminaries}

We begin with necessary background on importance sampling and the rendering equation.

\noindent{\bf Importance sampling}
Monte Carlo sampling is a commonly used method for evaluating integrals~\cite{hammersley2013monte}. The principal challenge lies in approximating the integral with fewer samples while achieving reduced variance. Given an integral $F = \int_{\Omega} f(x)\Diff{x}$, where $f(x)$ is the target integrand function over the domain $\Omega$, and $x \in \mathbb{R}^D$ is a sample, importance sampling is performed by:
\begin{equation} 
F = \int_{\Omega} \frac{f(x)}{q(x)} q(x) \Diff{x} = \mathbb{E} \left[ \frac{f(X)}{q(X)} \right] \approx \frac{1}{N} \sum_{i=1}^{N}\frac{f(X_i)}{q(X_i)},
\end{equation} 
where the proxy PDF $q(x)$ is designed to approximate the distribution of 
$f(x)$ in a way that minimizes the variance of the estimator.
Ideally, $q(x)$ should be chosen so that it matches the shape of 
$f(x)$ closely, meaning that areas where $f(x)$ has higher values are sampled more frequently.
This helps in achieving a more accurate estimate of the integral with fewer samples.

\noindent{\bf Rendering equation} The rendering equation~\cite{kajiya1986rendering} models the interaction between lighting and surfaces. Given the incident radiance $L_i(\boldsymbol\omega_i, \boldsymbol{x})$ and the material properties at a surface point $\boldsymbol{x}$ with normal vector $\boldsymbol{n}$, the rendering equation expresses the outgoing radiance in the following manner:
\begin{equation}
L_o(\boldsymbol{\omega}_o, \boldsymbol{x}) = \int_{\Omega} f(\boldsymbol{\omega}_o, \boldsymbol{\omega}_i, \boldsymbol{x}) L_i(\boldsymbol\omega_i, \boldsymbol{x}) (\boldsymbol\omega_i\cdot \boldsymbol{n}) d\boldsymbol\omega_i\,,
\label{eq:rendering_equation}
\end{equation}
where $f(\boldsymbol{\omega}_o, \boldsymbol{\omega}_i, \boldsymbol{x})$ is the Bidirectional Reflectance Distribution Function (BRDF). Here, we consider the microfacet BRDF parameterization~\cite{torrance1967theory}, including material properties such as albedo $\boldsymbol{a} \in [0,1]^3$, metallic $m \in [0,1]$, and roughness $r \in [0,1]$. Formally, this BRDF can be decomposed into a diffuse term $f_d$ and a specular term $f_s$ defined as:
\begin{equation}
f_d = \frac{(1 - m)\boldsymbol{a}}{\pi},
    \label{eq:f_d}
\end{equation}
\begin{equation}
f_s(\boldsymbol{\omega}_i,\boldsymbol{\omega}_o, \boldsymbol{x}) = \frac{D F G}{4 (\boldsymbol{\omega}_i \cdot \mathbf{n}) (\boldsymbol{\omega}_o \cdot \mathbf{n})},
    \label{eq:f_s}
\end{equation}
where $D$, $F$, and $G$ represent the normal distribution function, the Fresnel term, and the geometry term, respectively.

\section{TensoFLow}

\noindent{\bf Architecture overview}
Our \model{} is comprised of two stages:
(I) The first stage involves reconstructing the scene geometry. We follow TensoSDF~\cite{li2024tensosdf} and utilize a tensorial signed distance field combined with a roughness-aware training strategy to achieve robust geometry reconstruction.
(II) The second stage focuses on modeling the material and lighting, which is central to inverse rendering and constitutes the core of our approach, as detailed below.

\subsection{Material and lighting estimation}
\paragraph{Parametrization}
In \model{}, we represent the material of a scene 
with  a Vector-Matrix tensorial encoder~\cite{chen2022tensorf},
due to its high capacity and flexibility.
For a surface point $\boldsymbol{x}$, its tensorial feature vector is defined as
\begin{equation}
    V_m(\boldsymbol{x}) = \mathrm{v}^X_{m,k} \circ \mathrm{M}^{YZ}_{m,k} \oplus \mathrm{v}^Y_{m,k} \circ \mathrm{M}^{XZ}_{m,k} \oplus \mathrm{v}^Z_{m,k} \circ \mathrm{M}^{XY}_{m,k},
\label{eq:tensorial_material}
\end{equation}
where $\mathrm{v}_k$ and $\mathrm{M}_k$ represent the $k$-th vector and matrix along the corresponding axis. The operators $\circ$ and $\oplus$ denote element-wise multiplication and concatenation, respectively. After concatenating this feature vector with the positional encoding $\mathbf{p}$ of the position $\boldsymbol{x}$, it is processed by a tiny MLP $\Theta_m$ to produce the material properties: 
\begin{equation}
    \{\boldsymbol{a}, m, r\} = \Theta_m(V_m, \mathbf{p}),
\label{eq:material_parametrization}
\end{equation}
where $\boldsymbol{a}$, $m$, and $r$ denote albedo, metallic, and roughness, respectively.
For the incident light modeling, we decompose it into spatially-aware indirect lighting and direct lighting originating from infinity:
\begin{equation}
    L(\boldsymbol{\omega}_i, \boldsymbol{x}) = V(\boldsymbol{\omega}_i, \boldsymbol{x})L_\mathrm{dir}(\boldsymbol{\omega}_i)+L_\mathrm{ind}(\boldsymbol{\omega}_i, \boldsymbol{x}),
\label{eq:lighting_decompose}
\end{equation}
where the visibility term $V(\boldsymbol{\omega}_i, \boldsymbol{x})$ is determined via ray tracing on the extracted mesh. The indirect and direct lightings are modeled by a tiny MLP $\Theta_\mathrm{ind}$ and an environment map $\Theta_\mathrm{dir}$, respectively:
\begin{align}
    L_\mathrm{ind}(\boldsymbol{\omega}_i, \boldsymbol{x})&=\Theta_\mathrm{ind}(\mathbf{p}, \boldsymbol{\omega}_i),\\
    L_\mathrm{dir}(\boldsymbol{\omega}_i) &= \Theta_\mathrm{dir}(\boldsymbol{\omega}_i),
\label{eq:lighting_parametrization}
\end{align}

\paragraph{Rendering}
Given the reconstructed signed distance field from the first stage, the intersection point between each ray and the surface can be determined. The material properties and incident lighting at this intersection can then be queried as described earlier.
In accordance with \cref{eq:rendering_equation}, we employ Monte Carlo sampling to evaluate the integral in the rendering equation. Due to differing characteristics of the diffuse and specular terms, separate importance samplers are used to minimize variance in the Monte Carlo sampling process:
\begin{align}
    \boldsymbol{c}_{\rm diffuse} &= \frac{1}{N_{d}}\sum_i^{N_{d}} \frac{(1-m) \frac{\boldsymbol{a}}{\pi} L(\boldsymbol{\omega}_i, \boldsymbol{x}) (\boldsymbol{\omega}_i \cdot \boldsymbol{n})}{q_{\rm d}(\boldsymbol{\omega}_i; \boldsymbol{x})},\label{eq:diff_mc}\\
    \boldsymbol{c}_{\rm specular} &= \frac{1}{N_s} \sum_{i}^{N_{s}} \frac{\frac{D F G}{4 (\boldsymbol{\omega}_i \cdot \boldsymbol{n}) (\boldsymbol{\omega}_o \cdot \boldsymbol{n})} L(\boldsymbol{\omega}_i, \boldsymbol{x}) (\boldsymbol{\omega}_i \cdot \boldsymbol{n})}{q_{\rm s}(\boldsymbol{\omega}_i; \boldsymbol{x})},\label{eq:spe_mc}
\end{align}
where $q_{\rm d}(\boldsymbol{\omega}_i)$ and $q_{\rm s}(\boldsymbol{\omega}_i)$ are the PDFs of the importance samplers, which we model them using the proposed normalizing flow (Sec.~\ref{sec:flow_nis}). The final color is computed as: 
\begin{equation}
    \boldsymbol{c} = \boldsymbol{c}_{\rm diffuse} + \boldsymbol{c}_{\rm specular},\label{eq:final_color}
\end{equation}

\subsection{Tensorial flow-based importance sampling}\label{sec:flow_nis}

For evaluating the rendering equation,
prior works often employ pre-defined distributions as the importance samplers.
For instance, in NeRO~\cite{liu2023nero} a cosine-weighted distribution is used as the importance sampler for the diffuse term, and the GGX distribution~\cite{cook1982reflectance} for the specular term.
However, such samplers cannot closely match the shape of the target integrand 
with complex spatial and directional variations for a typical scene,
leading to high variance and performance degradation.

\begin{figure}[t]
    \centering
    \includegraphics[width=\linewidth]{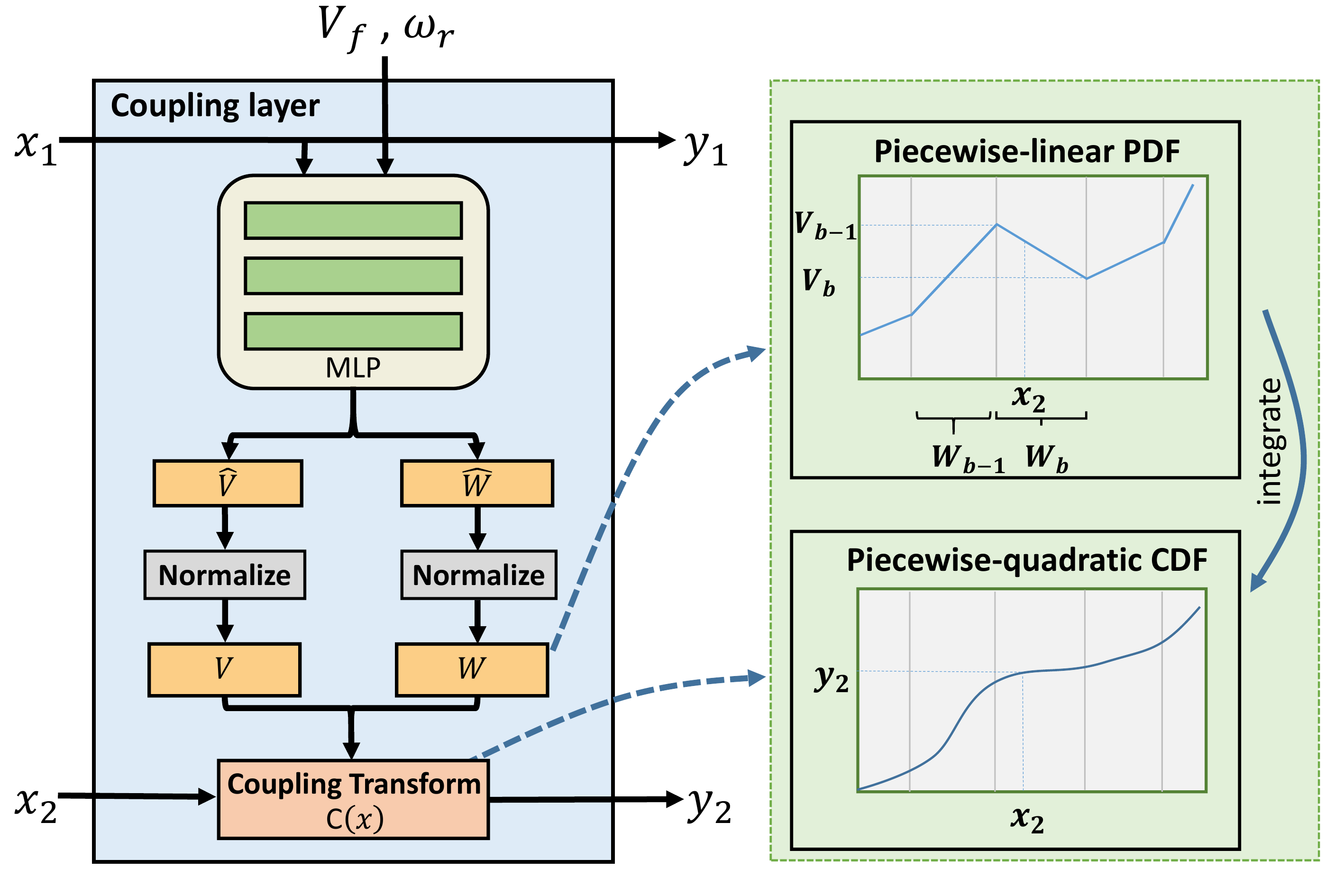}
    \caption{Illustration of a piecewise-quadratic coupling layer, incorporating tensorial latent feature $V_f$ and reflected direction $\boldsymbol{\omega}_r$ as spatial and directional priors.
}
\label{fig:coupling_layer}
\end{figure}

\subsubsection{Flow-based sampler}
To overcome this limitation, we propose learning the importance samplers.
The normalizing flow framework \cite{dinh2014nice} is adopted as the sampler learner due to its following theoretical advantage.
For a given direction $\boldsymbol{\omega}$ with two degrees of freedom in spherical coordinates, the distribution of the integrand function may exhibit inter-dependencies across different coordinate dimensions. We can express the variable $\boldsymbol{\omega}$ as a bijective transformation of an independent variable $\boldsymbol{z}$, which follows a uniform distribution $(0,1)^2$: $\boldsymbol{\omega} = h(\boldsymbol{z})$. Using the change-of-variable formula, the PDF of $\boldsymbol{\omega}$ can be expressed as:
\begin{equation} 
q(\boldsymbol{\omega}) = q(\boldsymbol{z}) \left| \det\left( \frac{\partial h(\boldsymbol{z})}{\partial \boldsymbol{z}^T} \right) \right|^{-1} ,
\label{eq:chang-of-variable} 
\end{equation}
where $q(\boldsymbol{z})$ represents the PDF of $\boldsymbol{z}$ in $\mathcal{U}(0,1)^2$, which is straightforward to evaluate.

However, the selection of the bijective mapping $h$ must satisfy certain conditions: it must be invertible, and it should allow efficient computation of the Jacobian determinant. To meet these requirements, we employ NICE~\cite{dinh2014nice}, a type of autoregressive flow that utilizes "coupling layers" to enable fast sampling and density evaluation. The final mapping $h$ is then formed as a composition of simpler bijective transformations: $h = h_n \circ \cdots \circ h_1$.

Given an input vector $x = (x_1, x_2) \in \mathbb{R}^2$ for a mapping $h_i$, the output vector $y = (y_1, y_2) \in \mathbb{R}^2$ is defined as:
\begin{equation}
y_1 = x_1 \quad , \quad y_2 = C\big(x_2; m(x_1)\big) \,,
\label{eq:coupling_layer}
\end{equation}
where the coupling transform $C:\mathbb{R} \leftrightarrow \mathbb{R}$ is a separable and invertible transformation, and $m(x)$, parameterized by a neural network, acts as the parameter of $C$. This structure allows the inverse of the coupling layer to be computed directly: 
\begin{equation}
x_1 = y_1 \quad , \quad x_2 = C^{-1}\big(y_2; m(y_1)\big) .
\label{eq:coupling_layer_inverse}
\end{equation}
The separability of $C$ ensures that the Jacobian matrix $J_i$ of $h_i$ is triangular, simplifying the computation of its determinant to the product of its diagonal elements. 

By applying the change-of-variable formula, the Jacobian determinant of the composite mapping $h$ is expressed as the product of the individual $J_i$: $\left|\det J\right| = \prod_i \left|\det J_i\right|$. We note that, with sufficiently expressive coupling transformations and adequate coupling layers, the sampling distribution $q(x)$ can be modeled as an arbitrarily complex function, allowing it to closely match the shape of any integrand in the rendering equation. Importantly, each adjacent coupling layer preserves different dimensions of the input vector (\cref{eq:coupling_layer}), ensuring the mapping's expressiveness across all the dimensions.

The normalizing flow defined by the above mapping acts as a distribution from which we can query the PDF or draw samples, corresponding to $h^{-1}$ and $h$, respectively. Our method supports two primary applications in the inverse rendering pipeline: inferring the PDF value for a given direction and sampling a direction along with its PDF value:
\begin{align}
\text{Inference:}&\quad \quad \quad q(\boldsymbol{\omega})\leftarrow h^{-1}(\boldsymbol{\omega}),\\
\text{Sampling:}&\quad (\boldsymbol{\omega},q(\boldsymbol{\omega}))\leftarrow h(\boldsymbol{u}), \boldsymbol{u} \sim \mathcal{U}(0, 1)^2,
\label{eq:flow_application}
\end{align}
Our experiments reveal that modeling the distribution of the half vector $\boldsymbol{\omega}_h$ yields better performance compared to directly modeling $\boldsymbol{\omega}_i$. By modeling the distribution $\boldsymbol{\omega}_h$, $q(\boldsymbol{\omega}_i)$ can be derived using the change-of-variable formula (\cref{eq:chang-of-variable}), enabling both PDF inference and sampling of $\boldsymbol{\omega}_i$.

\subsubsection{Tensorial coupling transformation}
Given the spatially-dependent nature of the integrand in the rendering equation, 
we further parameterize the scene space by incorporating an additional tensorial representation as:
\begin{equation}
    V_f(\boldsymbol{x}) = \mathrm{v}^X_{f,k} \circ \mathrm{M}^{YZ}_{f,k} \oplus \mathrm{v}^Y_{f,k} \circ \mathrm{M}^{XZ}_{f,k} \oplus \mathrm{v}^Z_{f,k} \circ \mathrm{M}^{XY}_{f,k}.
\label{eq:tensorial_flow}
\end{equation}
The latent feature $V_f$, together with the reflected direction $\boldsymbol{\omega}_r = 2(\boldsymbol{\omega}_o \cdot \boldsymbol{n})\boldsymbol{n} - \boldsymbol{\omega}_o$ for directional dependency, serves as conditions to guide the network $m_i$ within each coupling layer. This is crucial, as the shape of the integrand in the rendering equation is influenced by both the spatial position and the outgoing direction $\boldsymbol{\omega}_o$.

We consider the piecewise-quadratic coupling transform~\cite{muller2019neural} as our coupling transformation $C$
for its high expressiveness. Specifically, it achieves the bijective property by defining a Cumulative Distribution Function (CDF): $y = C(x; m(x';V_f, \boldsymbol{\omega}_r)) = \int_0^x p(t; m(x';V_f, \boldsymbol{\omega}_r)) \Diff t$, where $p$ is a spatially and directionally aware PDF derived from the output of network $m$. This results in the Jacobian determinant of the coupling layer being equal to $p(x)$. Here, the network $m$ takes the concatenation of $x'$, $V_f$, and $\boldsymbol{\omega}_r$ as input, generating $\widehat{V}$ and $\widehat{W}$ for subsequent use.
We then construct the piecewise linear PDF using  $K+1$ vertices, as illustrated in \cref{fig:coupling_layer}. It employs $K+1$ vertical coordinates $\widehat{V} \in \mathbb{R}^{K+1}$ and $K$ horizontal differences $\widehat{W} \in \mathbb{R}^K$, with $K$ as the predefined bin count for the distribution. Following softmax normalization, $W = \sigma(\widehat{W})$ represents the bin widths, while $V$ defines the PDF values at each vertex.
\begin{equation}
V_{i} =
\frac{\exp\left(\widehat{V}_{k}\right)}
{\sum_{k=1}^{K} \frac{\exp\left(\widehat{V}_{k}\right) + \exp\left(\widehat{V}_{k+1}\right)}{2} W_{k}} \,,
\end{equation}
Consequently, given an input $x$, the corresponding PDF value can be queried as
\begin{equation}
p(x) =\mathrm{lerp}{(V_{b},V_{b+1},{\alpha})}\,,
\label{eq:piecewise-linear-pdf}
\end{equation}
where $\alpha = (x - \sum_{k=1}^{b-1} W_{k}) / W_{k}$ defines the relative position of $x$ within the corresponding bin $b$. The piecewise-quadratic CDF can be computed by integrating the PDF: 
\begin{equation}
C(x) = \frac{\alpha W_{b}}{2}((2-\alpha)V_{b} + \alpha V_{b+1}) + \sum_{k=1}^{b - 1}
\frac{V_{k}+V_{k+1}}{2} W_{k} .
\end{equation}
We employ $N$ coupling layers with piecewise-quadratic transformation to construct the compound mapping $h$, ensuring it is sufficiently expressive for modeling the importance sampler. 
\begin{figure*}[ht]
    \centering
    \includegraphics[width=\linewidth]{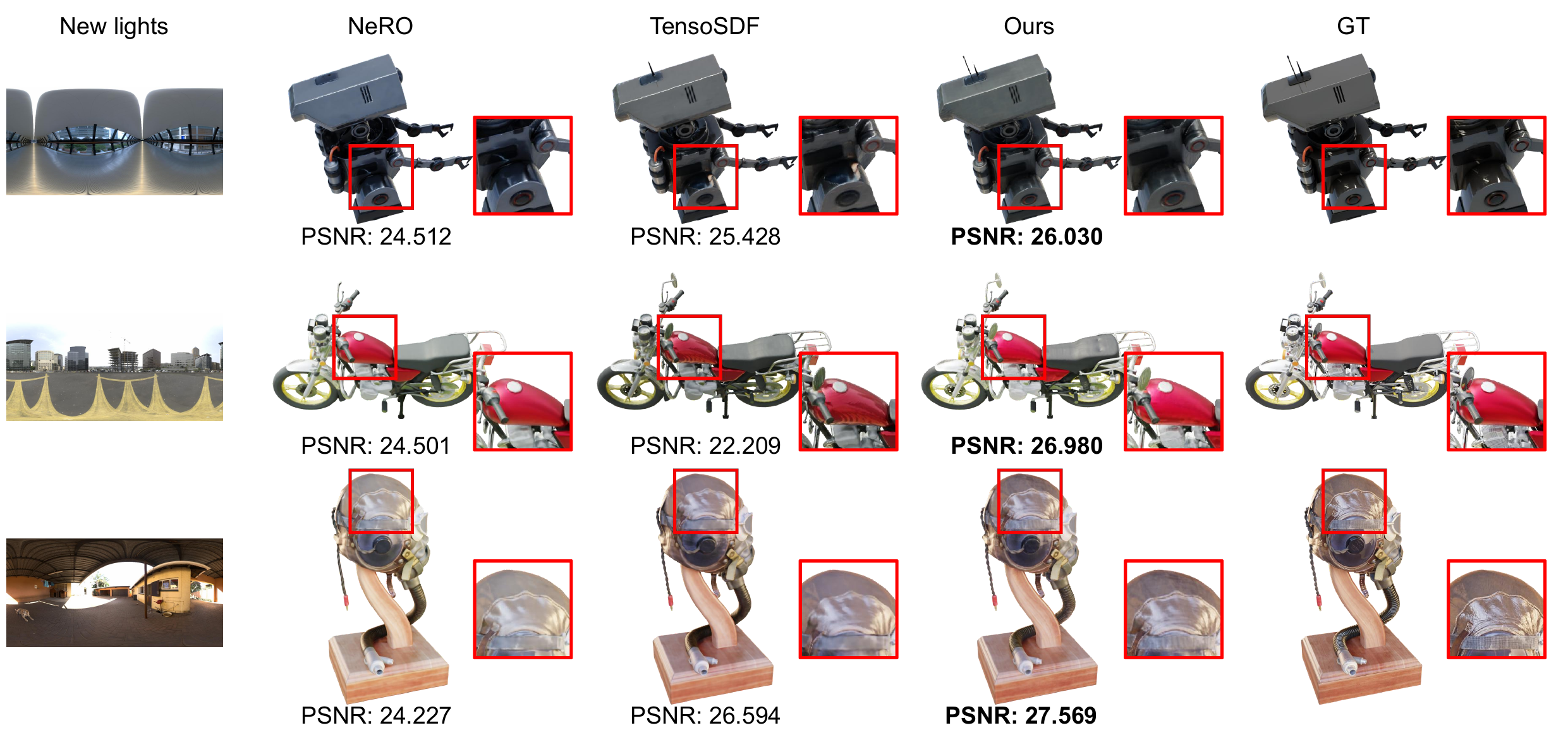}
    \caption{
    Qualitative comparison of relighting quality on TensoSDF dataset~\cite{li2024tensosdf}.
}
\label{fig:tensosdf_relighting}
\end{figure*}

\subsection{Training} \label{sec:training}
During model training, we optimize the tensorial encoder $V_f$ and the networks $\{m_i\}_{i=1...N}$ by minimizing the cross-entropy loss between the integrand of the rendering equation and the PDF of the incident directions inferred from the normalizing flow:
\begin{align}
    \mathcal{L}_\mathrm{ce} &= -\int_\Omega I(\boldsymbol{\omega}_i, \boldsymbol{\omega}_o, \boldsymbol{x}) \log{q(\boldsymbol{\omega}_i)} \Diff{\boldsymbol{\omega}_i} \\
    &= \mathbb{E}\left[-\frac{I(\boldsymbol{\omega}_i, \boldsymbol{\omega}_o, \boldsymbol{x})}{\hat{q}(\boldsymbol{\omega}_i)}\log{q(\boldsymbol{\omega}_i)}\right] \,,
    \label{eq:loss-entropy}
\end{align}
where $\hat{q}(\boldsymbol{\omega}_i)$ represents the PDF of the sampled incident directions from the current importance sampler, a frozen copy of the trainable normalizing flow updated every $N_\mathrm{update}$ iterations. $q(\boldsymbol{\omega}_i)$ denotes the PDF inferred from the current training normalizing flow, and $I(\boldsymbol{\omega}_i, \boldsymbol{\omega}_o, \boldsymbol{x})$ is the integrand function evaluated with the sampled incident direction.  We use different importance samplers for the diffuse and specular terms, resulting in separate losses: $\mathcal{L}_\mathrm{ce}^\mathrm{d}$ and $\mathcal{L}_\mathrm{ce}^\mathrm{s}$. 
The total training loss is thus defined as: 
\begin{equation}
    \mathcal{L} = \mathcal{L}_c + \lambda_m\mathcal{L}_m + \lambda_\mathrm{ce}^\mathrm{d}\mathcal{L}_\mathrm{ce}^\mathrm{d} + \lambda_\mathrm{ce}^\mathrm{s}\mathcal{L}_\mathrm{ce}^\mathrm{s},
\end{equation}
where $\mathcal{L}_c$ and $\mathcal{L}_m$ represent the RGB rendering loss and the material regularization loss~\cite{liu2023nero}, respectively.

\begin{figure*}[ht]
    \centering
    \includegraphics[width=\linewidth]{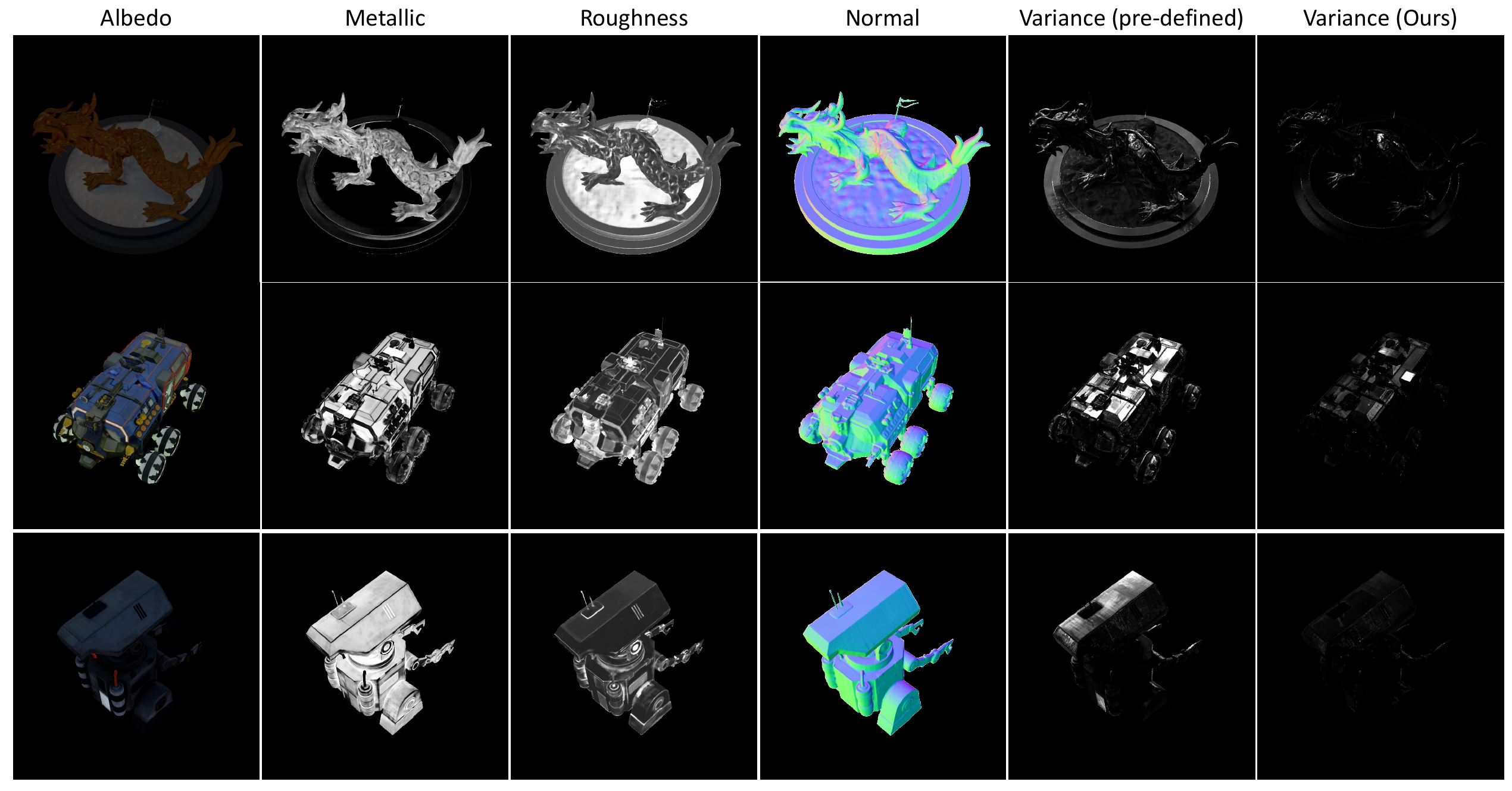}
    \caption{
    Visualization of material decomposition, normal map, and per-pixel variance in rendering equation evaluation, comparing results with a pre-defined sampler and our proposed tensorial normalizing flow. The per-pixel variance results are scaled by $1000$ for clearer visualization.
}
\label{fig:tensosdf_material}
\end{figure*}
\begin{table*}[ht]
    \setlength\tabcolsep{2pt}
    \centering
    \caption{
    Quantitative comparison of relighting quality on TensoSDF dataset~\cite{li2024tensosdf}.
    }
    \resizebox{0.95\linewidth}{!}{
    \begin{tabular}{c|ccccc|ccccc|ccccc}
    \hline\hline
    &   \multicolumn{5}{c|}{PSNR $\uparrow$} & \multicolumn{5}{c|}{SSIM $\uparrow$} & \multicolumn{5}{c}{LPIPS $\downarrow$}  \\
    & TensoIR~\cite{jin2023tensoir} & NeRO~\cite{liu2023nero} & NeiLF++~\cite{zhang2023neilf++} & TensoSDF~\cite{li2024tensosdf} & TensoFLow & TensoIR & NeRO & NeiLF++ & TensoSDF & TensoFLow & TensoIR & NeRO & NeiLF++ & TensoSDF & TensoFLow \\ \hline
    Rover & 24.000 & 24.015 & 23.774 & 26.754 & \textbf{26.936} & 0.918 & 0.914 & 0.911 & 0.935 & \textbf{0.937} & 0.0801 & 0.0693 & 0.0754 & 0.0593 & \textbf{0.0566} \\
    Dragon & 25.104 & 25.644 & 24.099 & 27.899 & \textbf{27.900} & 0.895 & 0.919 & 0.901 & 0.936 & \textbf{0.937} &  0.1302 & 0.0898 & 0.0988 & 0.0775 & \textbf{0.0756} \\
    Motor & 19.219 & 22.158 & 20.142 & 22.754 & \textbf{25.510} & 0.906 & 0.917 & 0.894 & 0.930 & \textbf{0.944} &  0.0821 &  0.0702 & 0.0870 & 0.0681 & \textbf{0.0556} \\
    Helmet & 25.140 & 22.587 & 24.001 & 28.126 & \textbf{28.385} &  0.901 & 0.881 & 0.906 & 0.934 & \textbf{0.937} &  0.1040 & 0.1079 & 0.1056 & 0.0770 & \textbf{0.0707} \\
    Robot & 26.031 & 23.194 & 22.696 & 26.242 & \textbf{27.416} &  0.928 & 0.913 & 0.915 & 0.940 & \textbf{0.943} &  0.0931 &  0.0755 & 0.0782 & 0.0613 & \textbf{0.0577} \\
    Compressor & 20.753 & 21.624 & 19.740 & 24.049 & \textbf{25.453} &  0.868 &  0.878 & 0.844 & 0.916 & \textbf{0.929} &  0.1038 & 0.1073 & 0.1286 & 0.0830 & \textbf{0.0789} \\
    \hline
    Average & 23.375 & 23.204 & 22.410 & 25.971 & \textbf{26.933} & 0.903 & 0.904 & 0.895 & 0.932 & \textbf{0.938} & 0.0989 &  0.0867 & 0.0956 & 0.0710 & \textbf{0.0659} \\
    \hline\hline
    \end{tabular}
    }
    \label{tab:tensosdf_relighting}
\end{table*}

\begin{figure*}[t]
    \centering
    \includegraphics[width=\linewidth]{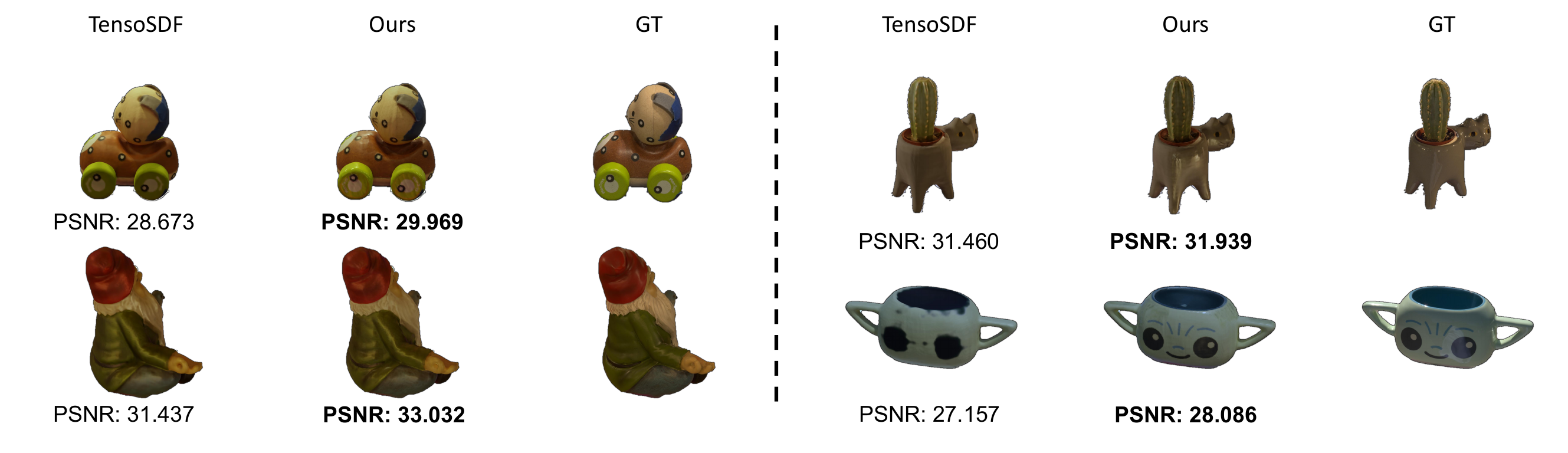}
    \caption{
    Qualitative comparison of relighted images with TensoSDF~\cite{li2024tensosdf} on the Stanford-ORB dataset~\cite{kuang2024stanford}.
}
\label{fig:orb_relighting}
\end{figure*}

\section{Experiment}

{\bf Datasets and metrics} 
We use the TensoSDF synthetic dataset~\cite{li2024tensosdf} and the real-world Stanford-ORB dataset~\cite{kuang2024stanford} for quantitative evaluation. To quantitatively evaluate the estimated material, we adopt three standard metrics for relighted images: PSNR, SSIM~\citep{wang2004image}, and LPIPS~\citep{zhang2018unreasonable}.

\noindent{\bf Competitors} 
We compare our approach with state-of-the-art NeRF-based methods selected for their different sampling strategies in evaluating the rendering equation: TensoIR~\cite{jin2023tensoir} and NeILF++\cite{zhang2023neilf++}, which utilize stratified sampling, and NeRO\cite{liu2023nero} and TensoSDF~\cite{li2024tensosdf}, which employ pre-defined importance sampling.

\noindent{\bf Implementation details} 
The training process consists of two stages. For the first stage, geometry reconstruction, we train the TensoSDF model for 180,000 iterations using the same settings as in TensoSDF~\cite{li2024tensosdf}. For the second stage, material and lighting estimation, we begin with a warm-up phase, employing cosine-weighted and GGX distributions as the importance samplers for the diffuse and specular components, respectively, over $N_\mathrm{warmup} = 1,000$ iterations. Once a coarse estimation of material and lighting is achieved at $N_\mathrm{ce} = 500$ iterations, we start optimizing the flow-related network by minimizing the cross-entropy losses $\mathcal{L}_\mathrm{ce}^\mathrm{d}$ and $\mathcal{L}_\mathrm{ce}^\mathrm{s}$. These flow-related networks include the tensorial encoder and the coupling layer network $\{m_i\}^d_{i=1...N}$ and $\{m_i\}^s_{i=1...N}$ for diffuse and specular respectively, each consists of a composition of $N=2$ piecewise-quadratic coupling layers. We model each $m_i$ using a three-layer tiny MLP with $64$ hidden unit. 
After the warm-up phase, the importance sampler is set as a frozen copy of the current normalizing flow, updated every $N_\mathrm{update} = 1,000$ iterations. After the warm-up phase, we employ the proposed tensorial importance sampler to sample incident directions. We sample 128 rays for the specular term. For the diffuse term, we apply the MIS~\cite{veach1995optimally} technique, sampling 64 rays from another normalizing flow and 512 rays from a cosine-weighted distribution. $\lambda_m$, $\lambda_\mathrm{ce}^\mathrm{d}$, and $\lambda_\mathrm{ce}^\mathrm{s}$ are set to $1.0$, $0.0001$, and $0.0001$, respectively. The training process takes 4 hours for the first stage and 2.5 hours for the second stage. All experiments are conducted on a single NVIDIA H100 GPU.

\subsection{Results on synthetic data}

We first perform experiments on the TensoSDF synthetic dataset~\cite{li2024tensosdf} to evaluate the effectiveness of our method in accurately reconstructing materials and lighting. Following the setups of NeRO~\cite{liu2023nero} and TensoSDF~\cite{li2024tensosdf}, we extract each object’s estimated material properties after optimization and use Blender to perform relighting based on the extracted mesh and material information.
As shown in \cref{tab:tensosdf_relighting}, our method consistently outperforms competitors across multiple scenes and metrics. This improvement highlights the benefit of our learnable importance sampler, which adapts to the integrand’s spatial and directional complexity, reducing Monte Carlo sampling variance and enhancing material and lighting estimation accuracy. Notably, NeRO underperforms relative to TensoSDF, due to its less accurate geometry reconstruction, which affects subsequent material and lighting estimations. In contrast, our method, built upon TensoSDF’s robust geometry, further refines material estimation with a learnable importance sampler, achieving a closer alignment with the integrand’s complex distribution in the rendering equation. 
In \cref{fig:tensosdf_relighting}, we provide a visual comparison of relighted images with NeRO and TensoSDF. Without a spatially and directionally aware importance sampler, the competitors rely on pre-defined sampling that struggles to match the shape of the integrand, leading to high variance and potential convergence to local minima during optimization.
\begin{table}
\centering
\small
\setlength\tabcolsep{3pt}
\caption{
Quantitative comparison of relighting quality on Stanford-ORB dataset~\cite{kuang2024stanford}.
}
\resizebox{0.95\linewidth}{!}{
\begin{tabular}{c|cc|cc|cc}
\hline\hline
& \multicolumn{2}{c|}{PSNR $\uparrow$} & \multicolumn{2}{c|}{SSIM $\uparrow$} & \multicolumn{2}{c}{LPIPS $\downarrow$} \\
& TensoSDF~\cite{li2024tensosdf} & TensoFLow & TensoSDF & TensoFLow & TensoSDF & TensoFLow \\
\hline
Teapot        & \textbf{29.971} & 29.342 &  \textbf{0.986} & \textbf{0.986} & 0.0276 & \textbf{0.0250}  \\
Gnome         & 28.488 & \textbf{29.179} & \textbf{0.956} & 0.960 & 0.0947 & \textbf{0.0899}  \\
Cactus        & 31.351 & \textbf{31.496} & 0.984 & \textbf{0.985} & \textbf{0.0362} & 0.0366  \\
Car           & 28.010 & \textbf{29.035} & 0.981 & \textbf{0.984} & 0.0367 & \textbf{0.0306}  \\
Grogu         & 30.595 & \textbf{30.735} & \textbf{0.990} & \textbf{0.990} & 0.0362 & \textbf{0.0385}  \\
\hline
Avg.          & 29.683 & \textbf{29.957} & 0.979 & \textbf{0.981} & 0.0463 & \textbf{0.0441}  \\
\hline\hline
\end{tabular}
}
\label{tab:orb_relighting}
\end{table}

In \cref{fig:tensosdf_material}, we visualize the estimated materials rendered by \model{} alongside per-pixel variance in Monte Carlo sampling of specular term. ``Variance (pre-defined)'' shows results using fixed importance samplers, specifically a GGX distribution. By contrast, our tensorial normalizing flow effectively models the importance sampler, allowing it to dynamically adjust to the integrand’s spatial and directional variations. This approach enables our method to produce detailed material maps—including albedo, metallic, and roughness, while significantly reducing variance with the same number of sampled rays.

\subsection{Results on real data}
We further evaluate our method on the real-world Stanford-ORB dataset~\cite{kuang2024stanford}, following the setup used in TensoSDF~\cite{li2024tensosdf}. Five objects from the dataset are selected for evaluation, and each object’s mesh and material properties are extracted from the optimized model and relight under two new environment lighting conditions. As shown in \cref{tab:orb_relighting}, our method demonstrates superior performance across most scenes. The performance increase, while notable, is less pronounced than on the TensoSDF synthetic dataset, likely due to the simpler geometry and lighting conditions of the objects in Stanford-ORB. In \cref{fig:orb_relighting}, we present a qualitative comparison between TensoSDF and our method, showing that our approach achieves more accurate material properties, which improves the quality of relighting.

\begin{figure}[t]
\centering
\begin{subfigure}[t]{\linewidth}
    \centering
    \includegraphics[width=\linewidth]{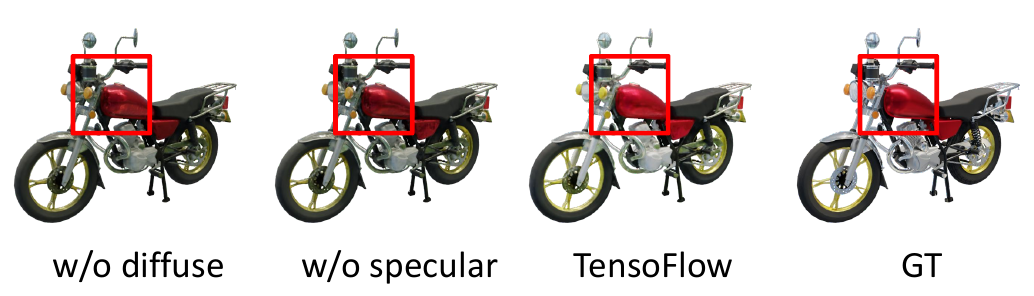}
    \caption{Ablation studies on learnable importance samplers.}
    \label{fig:ablation-importance}
\end{subfigure}
\begin{subfigure}[t]{\linewidth}
    \centering
    \includegraphics[width=\linewidth]{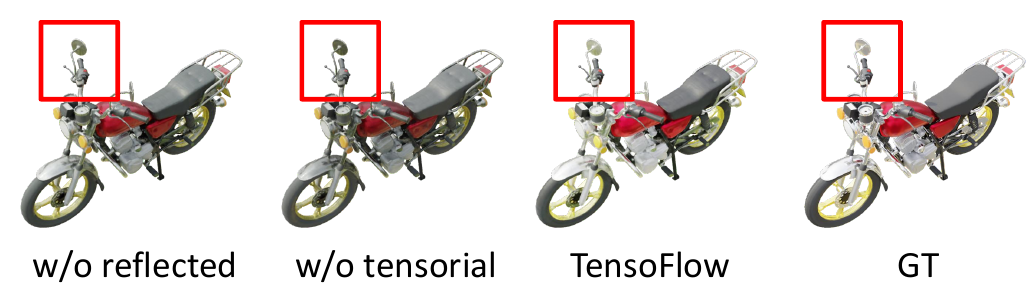}
    \caption{Ablation studies on spatial and directional priors.}
    \label{fig:ablation-prior}
\end{subfigure}
\begin{subfigure}[t]{\linewidth}
    \centering
    \includegraphics[width=\linewidth]{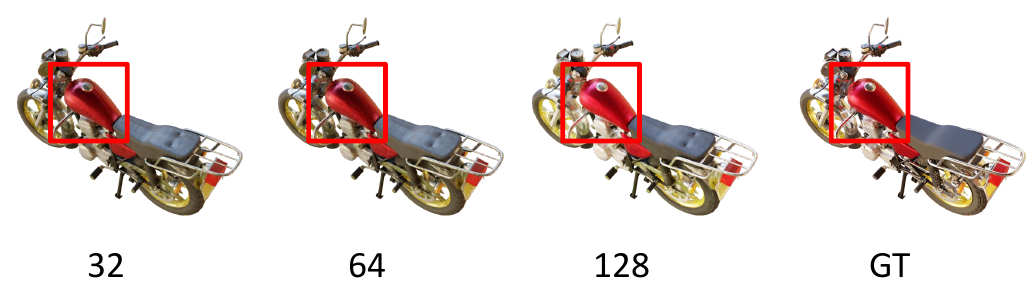}
    \caption{Ablation studies on number of sampled rays.}
    \label{fig:ablation-ray}
\end{subfigure}
\caption{
Ablation studies on various components of \model{}.
}
\vspace{-5mm}
\label{fig:ablation}
\end{figure}
\begin{table}[ht]
\centering
\small
\setlength\tabcolsep{10pt}
\caption{
Ablation studies on various components of \model{}.
}
\begin{tabular}{c|ccc}
\hline\hline
& PSNR $\uparrow$ & SSIM $\uparrow$ & LPIPS $\downarrow$ \\
\hline
w/o diffuse      & 26.433 & 0.937 & 0.0679  \\
w/o specular      & 26.203 & 0.936 & 0.0675  \\
One sampler      & 25.196 & 0.933 & 0.0736  \\
w/o half      & 26.474 & 0.937 &  0.0666 \\
\hline
w/o tensorial      & 26.058 & 0.932 & 0.0688  \\
w/o reflected      & 25.876 & 0.935  &  0.0703  \\
\hline
$N_s=32$    & 25.849 & 0.933 &  0.0706 \\
$N_s=64$      & 26.440 & 0.936 & 0.0674  \\
\hline
Full      & \textbf{26.933} & \textbf{0.938} & \textbf{0.0659} \\
\hline\hline
\end{tabular}
\vspace{-5mm}
\label{tab:ablation}
\end{table}
\subsection{Ablation}
As shown in \cref{tab:ablation}, we conduct ablation studies on various components of \model{} to evaluate their impact on relighting quality. We report the average metrics among all scenes in TensoSDF~\cite{li2024tensosdf} dataset.

\noindent{\bf Learnable importance sampler}
We investigate the impact of the learnable importance samplers in inverse rendering. In \cref{tab:ablation} and \cref{fig:ablation-importance}, “w/o diffuse” and “w/o specular” refer to using only the pre-defined importance samplers (cosine-weighted distribution for the diffuse term and GGX distribution for the specular term) for each component. “One sampler” indicates the use of a single normalizing flow to model both diffuse and specular terms simultaneously. The results demonstrate the importance of modeling diffuse and specular terms with separate normalizing flows for optimal performance. Additionally, “w/o half” represents modeling the distribution of the incident direction $\boldsymbol{\omega}_i$ directly instead of the half vector $\boldsymbol{\omega}_h$. The results show that modeling the distribution of the half vector $\boldsymbol{\omega}_h$ yields improved accuracy.

\noindent{\bf Spatial and directional prior}
In \cref{fig:ablation-prior}, we conduct ablation studies to assess the effectiveness of incorporating spatial and directional priors into the learnable importance sampler. In the settings “w/o tensorial” and “w/o reflected direction”, we exclude the tensorial latent feature $V_f$ and the reflected direction $\boldsymbol{\omega}_r$ as inputs to the network $m_i$ in the normalizing flow, respectively. The experimental results indicate that both inputs are essential for optimal performance.

\noindent{\bf Number of sampled rays}
In \cref{fig:ablation-ray}, we evaluate our method’s performance using varying numbers of sampled rays for evaluating the specular term in the rendering equation (\cref{eq:spe_mc}). Results indicate that with $N_s=64$ sampled rays, our method already achieves good performance, with further improvements as the sample count increases to $N_s=128$, yielding even greater accuracy.

\section{Conclusion}

In this paper, we have introduced \model{}, a novel approach for inverse rendering that utilizes a spatially and directionally aware, learnable importance sampler to tackle the complexity of light-surface interactions within the rendering equation. By parameterizing the sampler with normalizing flows, \model{} facilitates both efficient directional sampling of incident light and PDF inference. Our approach incorporates a learned tensorial representation along with the reflected direction to provide spatial and directional priors to the normalizing flow, resulting in a spatially and directionally aware sampling distribution. Additionally, we propose an optimization strategy to minimize the discrepancy between the integrand distribution and our learnable importance sampler. Extensive experiments on synthetic and real-world datasets demonstrate the effectiveness of \model{}, highlighting its superiority in material and lighting estimation compared to prior methods.

\paragraph{Limitation} Employing a normalizing flow as the importance sampler for the rendering equation slow down the sampling process, as each sampled direction requires a query to the flow.

\section*{Acknowledgments}
This work was supported in part by National Natural Science Foundation of China (Grant No. 62376060).

{
    \small
    \bibliographystyle{ieeenat_fullname}
    \bibliography{main}
}

\clearpage
\setcounter{page}{1}
\maketitlesupplementary

\section{More results}
We further provide a quantitative analysis of the sample variance in the rendering equation's evaluation. Specifically, the per-pixel variance is defined as:
\begin{equation}
    \mathrm{Var}=\frac{1}{N(N-1)}\sum_{i=0}^{N}\frac{(I_i-\overline{I})^2}{q_i},
\end{equation}
where $I$ denotes the integrand value of the rendering equation for a sampled incident direction, $\overline{I}$ represents the average integrand value across all samples, and $q_i$ is the probability density function (PDF) value for each sample. For quantitative evaluation, we report the mean variance across all pixels of the image for the specular term in \cref{tab:supp_variance}. Here, ``Pre-defined'' refers to the use of a pre-defined importance sampler (GGX distribution~\cite{walter2007microfacet}) for the rendering equation. The results clearly demonstrate that our learnable importance sampler achieves significantly lower variance compared to the predefined sampler. 

Furthermore, \cref{tab:supp_ablation_variance} shows the impact of each component of \model{} on variance reduction, highlighting the effectiveness of each module in improving the evaluation of the rendering equation.

\begin{figure*}[h]
    \centering
    \includegraphics[width=\linewidth]{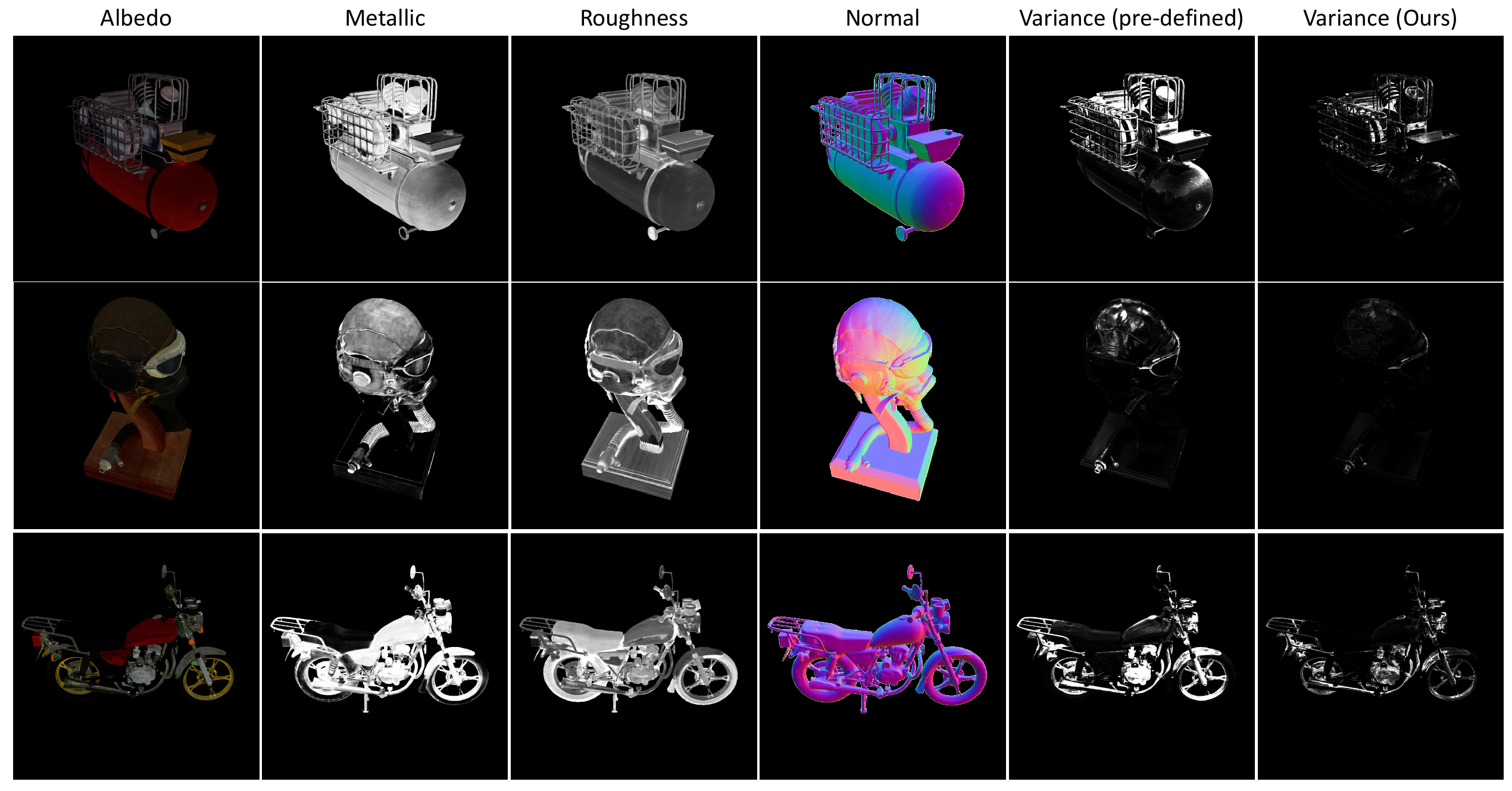}
    \caption{
    Visualization of material decomposition, normal map, and per-pixel variance in rendering equation evaluation, comparing results with a pre-defined sampler and our proposed tensorial normalizing flow. The per-pixel variance results are scaled by $1000$ for clearer visualization.
}
\label{fig:supp_tensosdf_material}
\end{figure*}

In \cref{fig:supp_tensosdf_material}, we provide additional visualizations of material decomposition, normal maps, and per-pixel variance on the TensoSDF dataset~\cite{li2024tensosdf}.

\begin{table}[ht]
\centering
\small
\setlength\tabcolsep{5pt}
\caption{
Quantitative evaluation of variance $\downarrow$ (in Units of $1e^{-5}$) in rendering equation evaluation using pre-defined variance and our learnable importance sampler.
}
\resizebox{0.6\linewidth}{!}{
\begin{tabular}{c|cc}
\hline\hline
 & Pre-defined & Ours \\
\hline
Rover       & 4.389  & \textbf{1.281} \\
Dragon      & 3.382  & \textbf{1.217} \\
Motor       & 11.957 & \textbf{3.340} \\
Helmet      & 2.357  & \textbf{0.691} \\
Robot       & 3.200  & \textbf{1.922} \\
Compressor  & 13.184 & \textbf{4.413} \\
\hline
Average     & 6.411  & \textbf{2.144} \\
\hline\hline
\end{tabular}
}
\label{tab:supp_variance}
\end{table}

\begin{table}[ht]
\centering
\setlength\tabcolsep{5pt}
\small
\caption{
Ablation studies on variance $\downarrow$ (in Units of $1e^{-5}$) across various components of \model{}.
}
\resizebox{0.6\linewidth}{!}{
\begin{tabular}{c|cc}
\hline\hline
         & Pre-defined & Ours \\
\hline
w/o half          & 4.955       & \textbf{1.773} \\
w/o tensorial     & 3.486       & \textbf{2.679} \\
w/o reflected     & 4.301       & \textbf{1.506} \\
$N_\mathrm{s}=32$ & 21.371      & \textbf{2.780} \\
$N_\mathrm{s}=64$ & 9.044       & \textbf{1.996} \\
Full              & 6.411       & \textbf{2.144} \\
\hline\hline
\end{tabular}
}
\label{tab:supp_ablation_variance}
\end{table}

\end{document}